\def\year{2019}\relax
\documentclass[letterpaper]{article}
\usepackage{aaai}
\usepackage{times}
\usepackage{helvet}
\usepackage{courier}
\usepackage{graphicx}
\usepackage{url}  %Required
\usepackage{hyperref}
\usepackage{booktabs}
\usepackage[table,xcdraw]{xcolor}
\begin{document}
\title{A Method to Facilitate Cancer Detection and Type Classification from Gene Expression Data using a Deep Autoencoder and Neural Network}
\author{Xi Chen\\
Department of Statistics\\
Department of Molecular and\\
Cellular Biochemistry\\
University of Kentucky\\
Lexington, KY 40356\\
\texttt{billchenxi@gmail.com}
\And
Jin Xie\\
Department of Statistics\\
University of Kentucky\\
Lexington, KY 40356\\
\texttt{jin.xie@uky.edu}
\And
Qingcong Yuan\\
Department of Statistics\\
Miami University\\
Oxford, OH 45056\\
\texttt{qingcong.yuan@miamioh.edu}
}
\maketitle

\begin{abstract}
\begin{quote}
With the increased affordability and availability of whole-genome sequencing, large-scale and high-throughput gene expression is widely used to characterize diseases, including cancers. However, establishing specificity in cancer diagnosis using gene expression data continues to pose challenges due to the high dimensionality and complexity of the data. Here we present models of deep learning (DL) and apply them to gene expression data for the diagnosis and categorization of cancer. In this study, we have developed two DL models using messenger ribonucleic acid (mRNA) datasets available from the Genomic Data Commons repository. Our models achieved 98\% accuracy in cancer detection, with false negative and false positive rates below 1.7\%. In our results, we demonstrated that 18 out of 32 cancer-typing classifications achieved more than 90\% accuracy. Due to the limitation of a small sample size (less than 50 observations), certain cancers could not achieve a higher accuracy in typing classification, but still achieved high accuracy for the cancer detection task. To validate our models, we compared them with traditional statistical models. The main advantage of our models over traditional cancer detection is the ability to use data from various cancer types to automatically form features to enhance the detection and diagnosis of a specific cancer type. 
\end{quote}
\end{abstract}

\section{Introduction}
Gene expression profiling technology has led to many significant biological discoveries. Much of the work built on gene expression data has focused on correlation between a small portion of the features from the data and various biological states and disease states, due to the high dimensionality and complexity of such data. With recent advances in computational power (parallel and massive computing on GPUs) and emergence of high-throughput data analysis and inference techniques, such as artificial intelligence, machine learning, and deep learning, fitting high-dimensional data is no longer the bottleneck of the in silico application of gene expression data. However, current solutions for disease detection, especially cancer detection methods, still rely on traditional feature selection and dimension reduction approaches, i.e. finding a small feature space of genes with expression values sufficiently correlated to the disease states. The assumption is that ignoring features it is assumed do not provide significant information, allows a reduction in computation complexity but still allows accurate single-class classification. 

This assumption misrepresents the mechanism of disease at system biological level. Models originating from this assumption are unscalable and more than not fail to generalize across different types of cancers without feature redesign. In addition, due to the small sample sizes available for individual cancers, these models cannot take full advantage of data from different cancer types. Even though many research studies have sought to conquer this issue, the majority still rely on principle component analysis, a linear statistical model approach. 

To deal with these problems and exceed the limits of the dimension-reduction paradigm in order to develop a more generalized model of cancer detection and typing classification, we propose in this paper a deep learning approach without the pains of feature space reduction/generation and that avoids the loss of potentially valuable information. We use an autoencoder method to learn high-level feature representations from the whole space of data features. In contrast to the previous methods, where one model is strictly appropriate for one type of cancer, our method allows typing classification of multiple cancers. In addition, this model does not rely on a dimension reduction approach; thus, even data collected from different microarray and RNA-sequencing platforms can still be used for detection and classification.

The remainder of this paper is organized as follows: Section 2 provides some background about gene expression and its data. Section 3 outlines the proposed method and several statistical models we used for comparison. Section 4 shows results of our method and compares them to other statistical learning models, and in Section 5 we discuss the results and future direction. Finally, Section 6 offers conclusions.

% ***********************************
\section{Gene Expression Data}
\subsection{Gene Expression Analysis}
Gene expression profiling has been adopted as the standard tool to measure the level of activity of genes within a given tissue. Gene expression data capture the amount of messenger ribonucleic acid (mRNA) synthesized during the cellular transcription, which ultimately will be translated into proteins. Thus, gene expression is fundamentally a direct measurement of the cellular functionality\cite{Lockhart:2000gs}. As cancers are a genetic and regulatory disease, cancer cells’ metabolic activities differ from those of normal cells, and this will be captured through gene expression data using microarray or RNA-seq\cite{Perou:1999fr}. Recently, more and more studies utilize a global gene-expression profiles of tumor cells and matched normal cells from the same origin to capture differences in expression. Though the number of human genes is about 22,000, gene expression data from the microarray or RNA-seq methods has a feature (column) number of 60,483, which includes extra transcriptome information—including both messenger or coding RNAs and a variety of non-coding RNAs (ncRNAs) \cite{Wang:2009di}.

In biomedical studies, gene expression analysis depends upon comparing relative proportions of reads; thus, data normalization is the essential and critical preprocessing step to correct unwanted biological effects, remove technical noise, and normalize differences between platforms. Three types of normalization are commonly used: reads per kilobase million (RPKM), fragment per kilobase million (FPKM) and transcripts per million (TPM). TPM is used in this paper to correct biases in gene length and sequencing depth because it allows comparison across different replicates \cite{Wagner:2012jf}. These values are generated through this pipeline by first aligning reads to the GRCh38 reference genome and then by quantifying the mapped reads. After the preprocessing pipeline, we obtain gene expression data as a matrix, which contains rows representing observations (sample) and columns representing features (corresponding to an array experiment), as shown in Figure 1.

\begin{figure}
\centering
\includegraphics[width=0.4\textwidth]{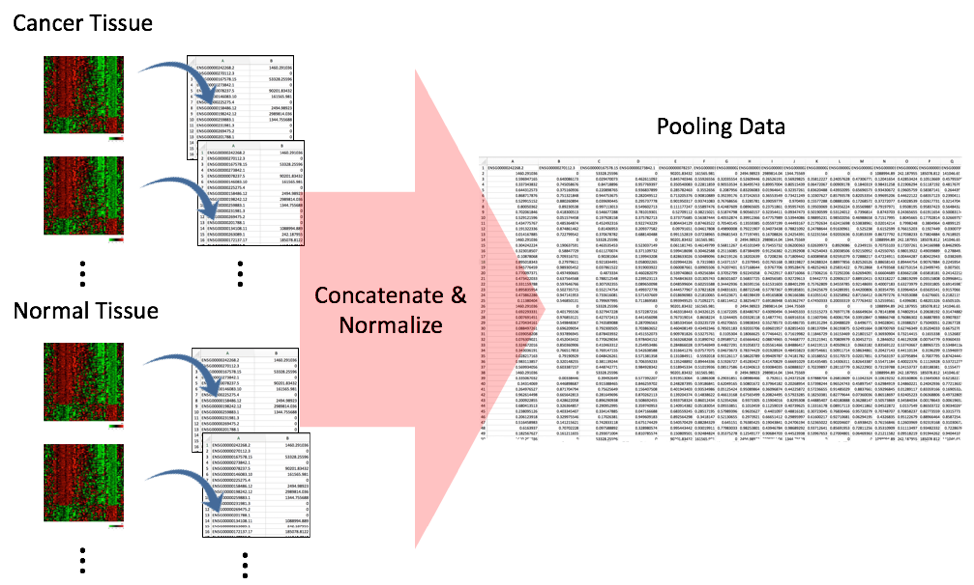}
\caption{Example gene expression data file.}
\end{figure}

\subsection{Gene Expression Datasets}
All gene expression data in this paper are from the Genomic Data Commons (GDC), a research program initiated by the National Cancer Institute (NCI) (\href{https://portal.gdc.cancer.gov}{The Genomic Data Commons Data Portal}). Data are collected from 29 major primary sites with 32 disease (cancer) types with gene expression data deposited at the GDC. These cancer types are KICH, LIHC, DLBC, OV, USC, LGG, THCA, ACC, LUAD, HNSC, BLCA, MESO, ESCA, UVM. CESC, LUSC, TGCT, PAAD, SARC, KIRP, UCEX, STAD, PCPG, KIRC, SKCM, THYM, PRAD, READ, GBM, BRCA, CHOL, and COAD.

% ***********************************
\section{Related Work}
Many researchers have proposed detecting cancer using gene expression data. The most fundamental approach is to use dimension reduction methods, such as PCA, to conduct feature selection, then apply traditional statistical classification methods\cite{Nguyen:2002ex}. Other feature selection approaches include the use of recursive feature elimination approaches in combination with other filter steps, e.g., (1) using univariate association filtering for feature selection\cite{Aliferis:to} and (2) applying SVM to find a small subset of gene expressions as a new and smaller feature space\cite{Ramaswamy:2001hc} that can achieve a similar result. Want et al. has shown a correlation-based selector combined with machine learning classification methods\cite{Hall:1999vl,Wang:2005bg} can achieve accuracy no worse than that achieved by other methods. A divide-and-conquer approach can also be used for feature selection: first, divide genes into smaller spaces, then, from these smaller subset, select the top-r informative genes, and later merge them into an informative subset of genes \cite{Sharma:2012ic}. The ensemble approach, which also combines different feature dimension reduction methods, must be mentioned as well\cite{Nanni:vj}. 

The majority of the methods mentioned above focus on reducing the gene expression dimension, or in a more general perspective, the feature generation (here equal to selection), based on certain criteria. One problem with these approaches is the limited scalability and generality of the classification models. Since features are designed, or selected, through a manual approach, it is almost impossible to apply the same features to new datasets or different classification tasks without re-designing, or re-selecting, the features. Because of this, models built on these features can neither be used to classify different types of cancers, or take advantage of the datasets from those cancers, since the selected features are highly correlated to the cancer from which they were generated\cite{Fakoor:2013tp}. The Pan-Cancer Epigenetic Biomarker Selection from Blood Samples\cite{XiChen} demonstrated that multiple cancer classification can be conduct through analysis of DNA methylation profiles with accuracy similar to that achieved with single classification. This suggests that gene expression profiles can be applied in a similar fashion by pooling all data and classifying multiple cancer types.

% ***********************************
\section{Methods}
In this section, we first introduce the computational configuration, then present the deep learning algorithms used to and explain a few keep DL techniques used to train our convolution and deep autoencoder model. Finally, we introduce several common statistical learning methods that we used for purposes of comparison with our new deep learning approaches. 

\subsection{Computational Architecture and Dataset Configuration}
To speed up the training process, our hardware configuration included a Nvidia Tesla K80 24GB graphics card with a 28 core Intel Xeon E5 v3 CPU. Python 3 was used with the TensorFlow machine learning framework, which wrapped in the high-level Python neural networks library Keras. Other Python libraries used included Numpy, Pandas, and matlibplot. The hardware and software setup is summarized in Figure 2.

\begin{figure}
\centering
\includegraphics[width=0.4\textwidth]{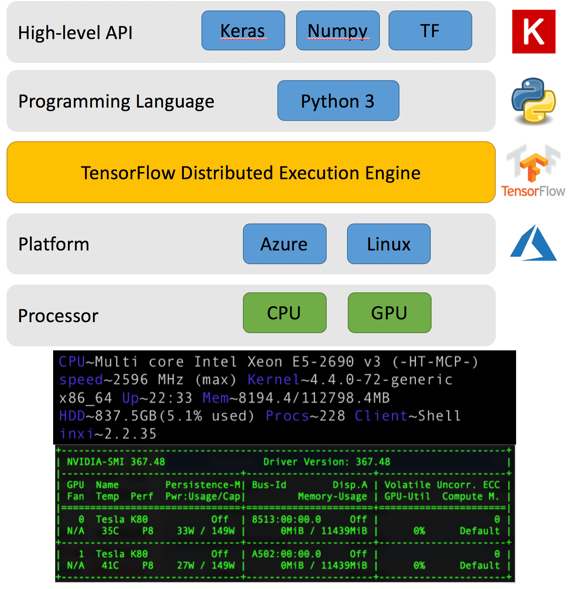}
\caption{Computational configuration.}
\end{figure}

We split the full dataset, using 80\% as a training set and reserving 20\% as a testing set after separately randomly shuffling the data for all cancer and normal datasets—they were shuffled separately because of an issue with unbalanced data between the two sets.

\subsection{Unsupervised learning via deep autoencoder}
Instead of manually selecting features or reducing the size of features, in the deep autoencoder approach\cite{Bengio:vb,Coates:wo,Ng:2013ur}, the model learns the feature representations through coupling encoder layers symmetrically with decoder layers (Figure 3). Both encoder and decoder layers are composed of a few shallow layers of deep-belief networks. The building block of the deep-belief networks are restricted Boltzmann machines\cite{and:tp}.

\begin{figure}
\centering
\includegraphics[width=0.4\textwidth]{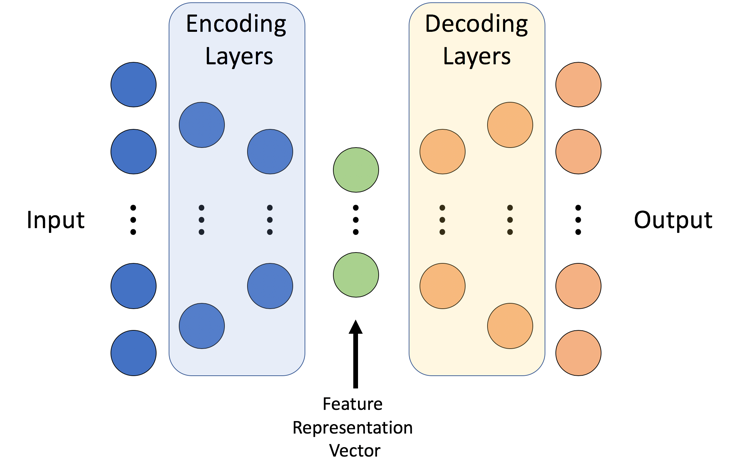}
\caption{Deep autoencoder structure.}
\end{figure}

In our approach, we have used five hidden layers with two encoder layers of 100 and 50 units each, a middle feature representation vector of 25 units, and two decoding layers with 50 and 100 units each. After training for 1000 epochs on the training data, we extracted the encoder layers and transformed them to a fully connected neural network with multi-classification as its output (Figure 4). Traditionally, the first encoding layer will have a higher number of units (nodes) than the input layer; however, since the number of our input layer is already greater than 60,000, we decided to use 100 units instead, the same as for the last decoding layer. 

The deep autoencoder essentially tries to learn a function $h_{(W,b)}(X) \approx X$, where $X$ is input, in this case, the gene expression data. In other words, it is trying to approximate the identity function with output $\hat{X}$ that is close to $X$\cite{Vincent:2010vu}. Through training, the model will learn the best compressed feature representation vector that can be decoded, or reconstructed, back into the original input. In this case, it eventually achieved a low-dimensional representation very similar to that achieved with PCA, but unlike PCA, the autoencoder is a nonlinear approach. The resulting feature representation vector captured the non-linearity of the relationship between expression of different genes.

The idea of fit the $X$ here we represent as normalized input that scaled into $[0,1]$ to the deep encoder model and output a result $\hat{X}$ similar to the input $X$, is to minimize the cross entropy between the input and output of the deep autoencoder:
$$ L(X,\hat{X})=-\sum_{k=1}^n [x_k \log⁡{\hat{x}_k}+(1-x_k)\log⁡{1-\hat{x}_k}]  $$

The output $\hat{X}$ is from the formula:
$$ \hat{X}=h_({W,b}(X)=\sigma(W^T \sigma(Wx+b_{encode} )^{\{l\}} +b_{decode} )^{\{l\}}  $$
where $\sigma(s) = \frac{1}{1+e^{-s}}$, $W\in{\rm I\!R}^{n\times m}$ ($l$ is the number of encoder/decoder layer, n the sample dimension, and m the number of neurons in each layer), $b_{en/de-coder, l} \in {\rm I\!R}$.

\subsection{Supervised transform learning with a neural network}
In order to perform the cancer detection and type classification, we froze the encoder layers learned from the deep autoencoder step and transformed them to a fully connected neural network, as shown in Figure 4.  The fully connected neural network is formed by stacking layers of neurons together. It includes three types of layers: an input layer, which is the output from the encoding layers of the deep autoencoder model, hidden layers, and an output layer. The neural network learns by adjusting weights through forward and backward propagation. Each neuron, or node, in the hidden layers uses a ReLU activation to achieve nonlinearity. Using labels provided by the data allows supervised learning to train the classifier. For the single-class classification, we used a logistic classifier, and for multiple-class classification, we chose a SoftMax classifier. 

\begin{figure}
\centering
\includegraphics[width=0.4\textwidth]{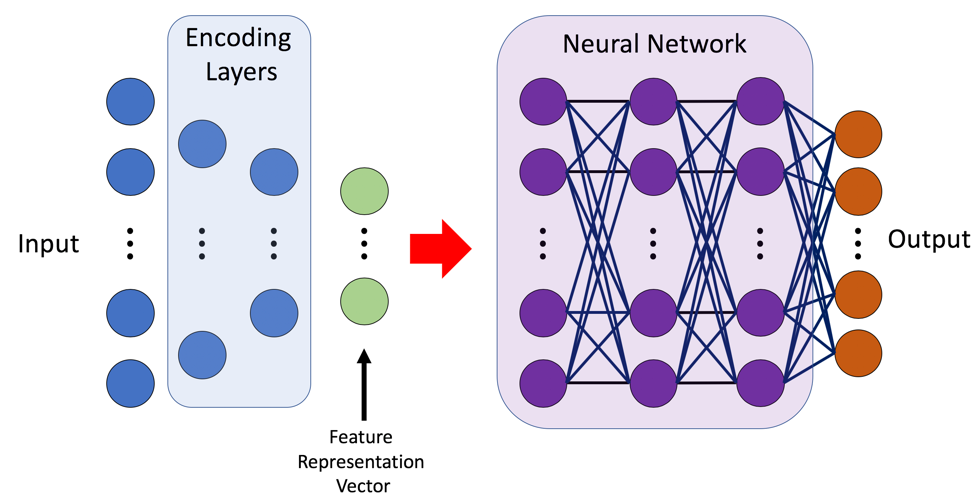}
\caption{Transform learning: using the feature representation vector from the autoencoder as the input of the multi-classification neural network.}
\end{figure}

\subsection{Machine and Statistical Learning with PCA} 
For purposes of comparison, we also employed other machine and statistical learning techniques to analyze the gene expression data. Similarly, normalized and shuffled training data were first put through a dimension reduction phase, as shown in Figure 5. The resulting extracted features are simply a linear function of the original input data, which lost all non-linearity of the relations between expressions of different genes\cite{Raina:2007gd}. Using the top 40 principle components from the dimension reduction, we further fit five machine and statistical learning models: linear discriminant analysis (LDA), neural network, K-nearest neighbor, random forest, and extremely randomized tree. 

\begin{figure}
\centering
\includegraphics[width=0.4\textwidth]{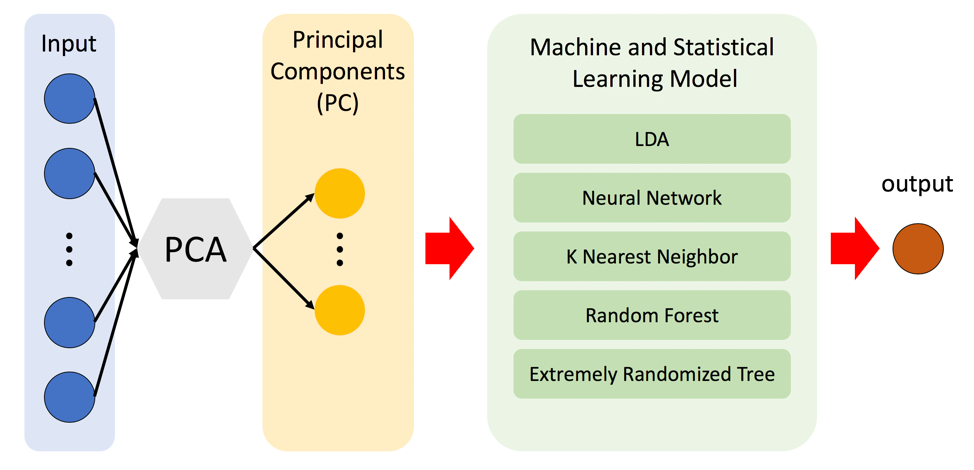}
\caption{Machine and Statistical Learning with PCA.}
\end{figure}

\section{Results and Discussion}
To demonstrate the feasibility and applicability of our proposed method we show the classification results for the testing data set (20\% of the total data). Since all comparison models that use PCA for dimension reduction can only apply to a single classification task, we tested our proposed method against comparison models only for the lung cancer data—since the lung cancer dataset is the only one with balanced sample sizes between the cancer and normal labels. The first three principle components from the PCA are visualized in Figure 6 and the classification comparison results are shown in Figure 7. The first three principle components illustrated in Figure 6 capture the most variate data and should, in theory, exhibit a clear separation between normal and cancer labels; however, almost all the data points mingled together at the bottom left of the figure. This suggests two important points: first, the linear function of the PCA is probably unable to capture the non-linearity of the data, and second, the classification power could not improve for the rest of the components, which have much lower eigenvalues, and so capture less variation.

\begin{figure}
\centering
\includegraphics[width=0.4\textwidth]{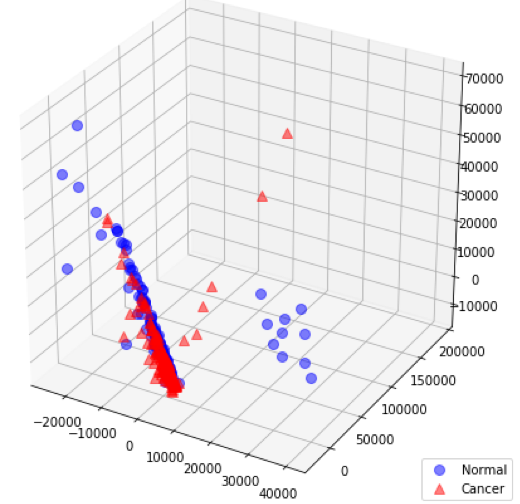}
\caption{First three principle components for lung cancer training dataset.}
\end{figure}

% \usepackage{booktabs}
\begin{table}
\centering
\caption{Classification results from the different machine and statistical learning models.}
\label{my-label}
\begin{tabular}{@{}cccc@{}}
\toprule
\rowcolor[HTML]{C0C0C0} 
Method & Accuracy & FPR & FNR \\ \midrule
\rowcolor[HTML]{EFEFEF} 
PCA-LDA & 98.30\% & 3.42\% & 0.48\% \\
\begin{tabular}[c]{@{}c@{}}PCA-Neural \\ Network\end{tabular} & 97.17\% & 3.42\% & 2.42\% \\
\rowcolor[HTML]{EFEFEF} 
\begin{tabular}[c]{@{}c@{}}PCA-K Nearest \\ Neighbor\end{tabular} & 95.75\% & 3.42\% & 4.83\% \\
\begin{tabular}[c]{@{}c@{}}PCA-Random \\ Forest\end{tabular} & 96.60\% & 6.16\% & 1.45\% \\
\rowcolor[HTML]{EFEFEF} 
\begin{tabular}[c]{@{}c@{}}PCA-Extremely \\ Randomized Tree\end{tabular} & 97.17\% & 4.79\% & 1.45\% \\ \bottomrule
\end{tabular}
\end{table}

\begin{figure}
\centering
\includegraphics[width=0.4\textwidth]{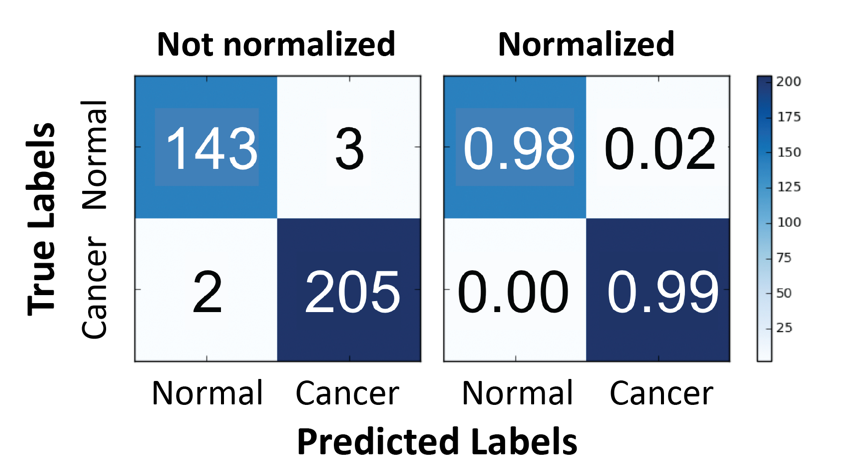}
\caption{Classification results for the proposed deep autoencoder approach.}
\end{figure}

As shown in Table 1, we achieved quite high accuracy using the PCA approach with different classification modes. The best model was the PCA + LDA, with 98.30\% accuracy; however, this is still not as good as the result from our proposed deep autoencoder approach. In addition, all PCA models had a high false positive rate (FPR) and false negative rate (FNR). For disease detection, the focus is on low FNR, especially since we do not want to predict a negative report when patient has cancer and thereby delay necessary treatment. Figure 7 shows the classifications from the deep autoencoder + neural network model. On the left of the figure is the confusion matrix without normalization—there, the values are counts of cases—and on the right are the normalized results as percentages. The results show our approach outperforms all the PCA models, with both high accuracy and very low FNR and FPR. 

By applying the multi-classification approach, we were able to construct an overall model for classifying 33 disease types (32 cancers + normal). In Figure 8, the background color, ranging from dark blue to white, represents the abundance of the data. In the perfect scenario, all diagonal elements of the confusion matrix should be 1, representing 100\% correction. In our case, the majority of the diagonal elements have high values and low FNR (suggest from the values in the non-diagonal elements). However, three cancers, LGG, UVM, and LUSC, all have a high misclassification rate, and GBM has a 100\% false negative rate.

\begin{figure}
\centering
\includegraphics[width=0.4\textwidth]{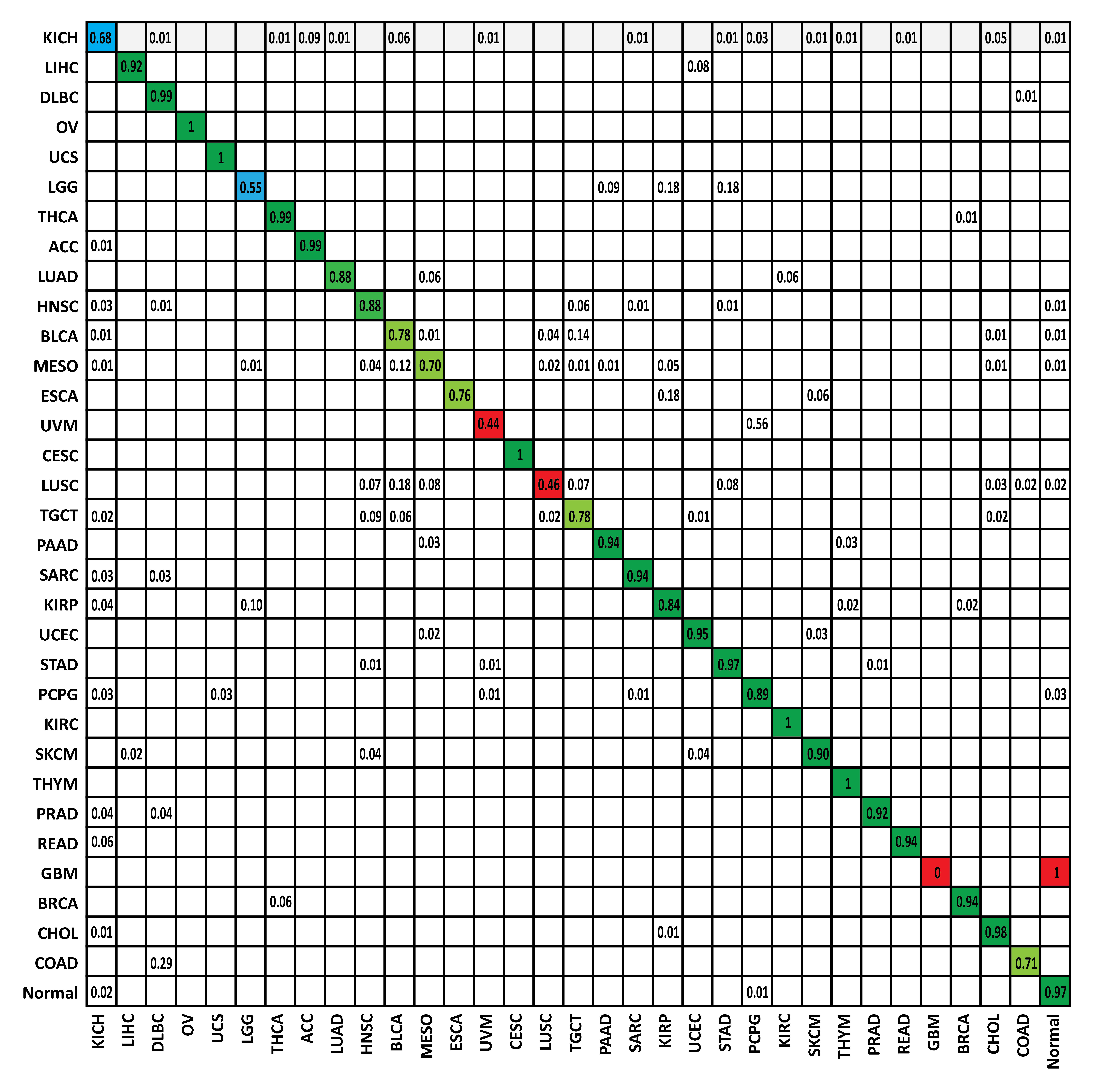}
\caption{Multi-classification results of the proposed autoencoder model.}
\end{figure}

All four problematic cancer types (LGG, low-grade glioma; UVM, uveal melanoma; LUSC, lung squamous cell carcinoma; and GBM, glioblastoma multiforme) had a very low sample abundance, with sample sizes much below 100 (some less than 50). This suggest the classification problem could root from the model being unable to “learn”, or extract, useful information from the data. Secondly, three of these cancers (LGG, UVM and GBM) are neural diseases. The metabolism of neural tissue is different than that of other tissues; therefore, the gene expression data might unable to capture the full pattern of the cancer. LUSM is the only one misclassified to many types with a low FPR; it occurs when abnormal lung cells multiple out of control and eventually spread (metastasize) to other tissues. This information helps explain that these type of cancer cells need to “disguise” themselves as other tissue types to metastasize; therefore, the gene expression pattern might be similar to the corresponding tissue type, causing it to be easily misclassified.

\section{Conclusion} 
In this paper, we propose a method to facilitate cancer detection and type classification from gene expression data using a deep autoencoder and neural network. Unlike traditional feature selection approaches, the method detailed here uses an autoencoder to automatically generate feature representations, thus addressing the very high dimensionality of gene expression data. This extracted feature vector captures the non-linearity of the data, is scalable for new data after training, and is able to generalize in multi-classification of different types of cancer. The results show that for cancer detection, a single-classification task, the proposed model can achieve higher accuracy and lower false positive and negative rate than traditional algorithms. For cancer type classification, a multi-classification task, it performs very well when sufficient sample data are available to train the model.

\section{ Acknowledgments}
The authors would also like to thank the anonymous referees for their valuable comments and helpful suggestions.

\bibliographystyle{aaai} \bibliography{ref}
\end{document}